\documentclass[10pt,twocolumn,letterpaper]{article}

\usepackage[pagenumbers]{iccv} 

\definecolor{iccvblue}{rgb}{0.21,0.49,0.74}
\usepackage[breaklinks,colorlinks,allcolors=iccvblue]{hyperref}
\usepackage{amsmath}
\usepackage{amssymb}
\usepackage{booktabs}
\usepackage{multirow} 
\usepackage{float}
\usepackage{makecell}
\usepackage{color}

\def\paperID{} 
\def\confName{ICCV}
\def\confYear{2025}

\title{Salvaging the Overlooked: Leveraging Class-Aware Contrastive Learning for Multi-Class Anomaly Detection}

\author{%
	Lei Fan$^{1}$\thanks{Corresponding author: \tt lei.fan1@unsw.edu.au}   \hspace{2em}  Junjie Huang$^{2}$ \hspace{2em}  Donglin Di$^{2}$ \hspace{2em} Anyang Su$^{2}$ \\  \hspace{2em}  Tianyou Song$^{3}$ \hspace{2em} Maurice Pagnucco$^{1}$ \hspace{2em} Yang Song$^{1}$ \\ %
	\vspace{-1em}
    \\
	$^{1}$UNSW Sydney \hspace{1em} $^{2}$DZ-Matrix \hspace{1em} $^{3}$Columbia University\\
	\tt \small \url{https://lgc-ad.github.io/}
    \vspace{-1em}
}

\def\etal{\emph{et al. }}
\def\ie{\emph{i.e., }}
\def\eg{\emph{e.g., }}

\begin{document}
\maketitle

\begin{abstract}
For anomaly detection (AD), early approaches often train separate models for individual classes, yielding high performance but posing challenges in scalability and resource management. Recent efforts have shifted toward training a single model capable of handling multiple classes. However, directly extending early AD methods to multi-class settings often results in degraded performance. In this paper, we investigate this performance degradation observed in reconstruction-based methods, identifying the key issue: inter-class confusion. This confusion emerges when a model trained in multi-class scenarios incorrectly reconstructs samples from one class as those of another, thereby exacerbating reconstruction errors. To this end, we propose a simple yet effective modification, called class-aware contrastive learning (CCL). By explicitly leveraging raw object category information (\eg carpet or wood) as supervised signals, we introduce local CL to refine multiscale dense features, and global CL to obtain more compact feature representations of normal patterns, thereby effectively adapting the models to multi-class settings. Experiments across five datasets validate the effectiveness of our approach, demonstrating significant improvements and superior performance compared to state-of-the-art methods. Notably, ablation studies indicate that pseudo-class labels can achieve comparable performance.

\end{abstract}

\section{Introduction}

Unsupervised Anomaly Detection (UAD), which involves training models using only normal samples to identify anomalous samples, has gained significant attention across various fields \cite{pang2021deep,fan2022grainspace,cheng2025open}. Existing studies tackle this unsupervised task by designing pretext tasks to transform it into a supervised problem, \eg reconstruction-based \cite{rd4ad,liu2023diversity}, synthesis-based \cite{li2021cutpaste,zavrtanik2021draem} methods or by using statistical models, \eg multivariate Gaussian distribution \cite{defard2021padim}, normalizing flows \cite{gudovskiy2022cflow} to estimate the patterns of normal samples.

\begin{figure}[t]
    \begin{subfigure}[b]{0.48\textwidth}
    \centering
    \renewcommand{\arraystretch}{0.9} 
    \setlength{\tabcolsep}{3pt} 
    \resizebox{\textwidth}{!}{ 
        \begin{tabular}{cccccc}
            \toprule
            \multirow{2}[0]{*}{Model} & \multirow{2}[0]{*}{One-for-one} & \multicolumn{4}{c}{One-for-all training strategies} \\
            \cmidrule(lr){3-6}
                  &       & \textit{Sequential} & \textit{Continual} & \textit{Joint} & \textit{CCL (our)} \\
            \cmidrule{1-6}
            RD \cite{rd4ad} & 94.0/97.2 & 74.9/91.3 &  76.9/91.4 & 90.3/96.9  & \textbf{94.6/98.3} \\
            DeSTSeg \cite{zhang2023destseg} & 93.8/94.7 &  64.5/76.3 & 68.5/78.0 & 90.7/90.9 & \textbf{92.5/92.3} \\
            \bottomrule
        \end{tabular}%
    }
    \caption{Evaluation of one-for-one models enhanced through four one-for-all training strategies: \textit{Sequential}, \textit{Continual}, \textit{Joint}, and CCL. Results are reported as average I- / P-AUROC (\%) across four datasets \cite{visa_ad,mvtec_ad,real_ad,btad_ad}. Detailed results are provided in the \textit{Supp.} }
    
    \end{subfigure}
    \begin{subfigure}[b]{0.48\textwidth}
        \centering        
        \includegraphics[width=1\textwidth]{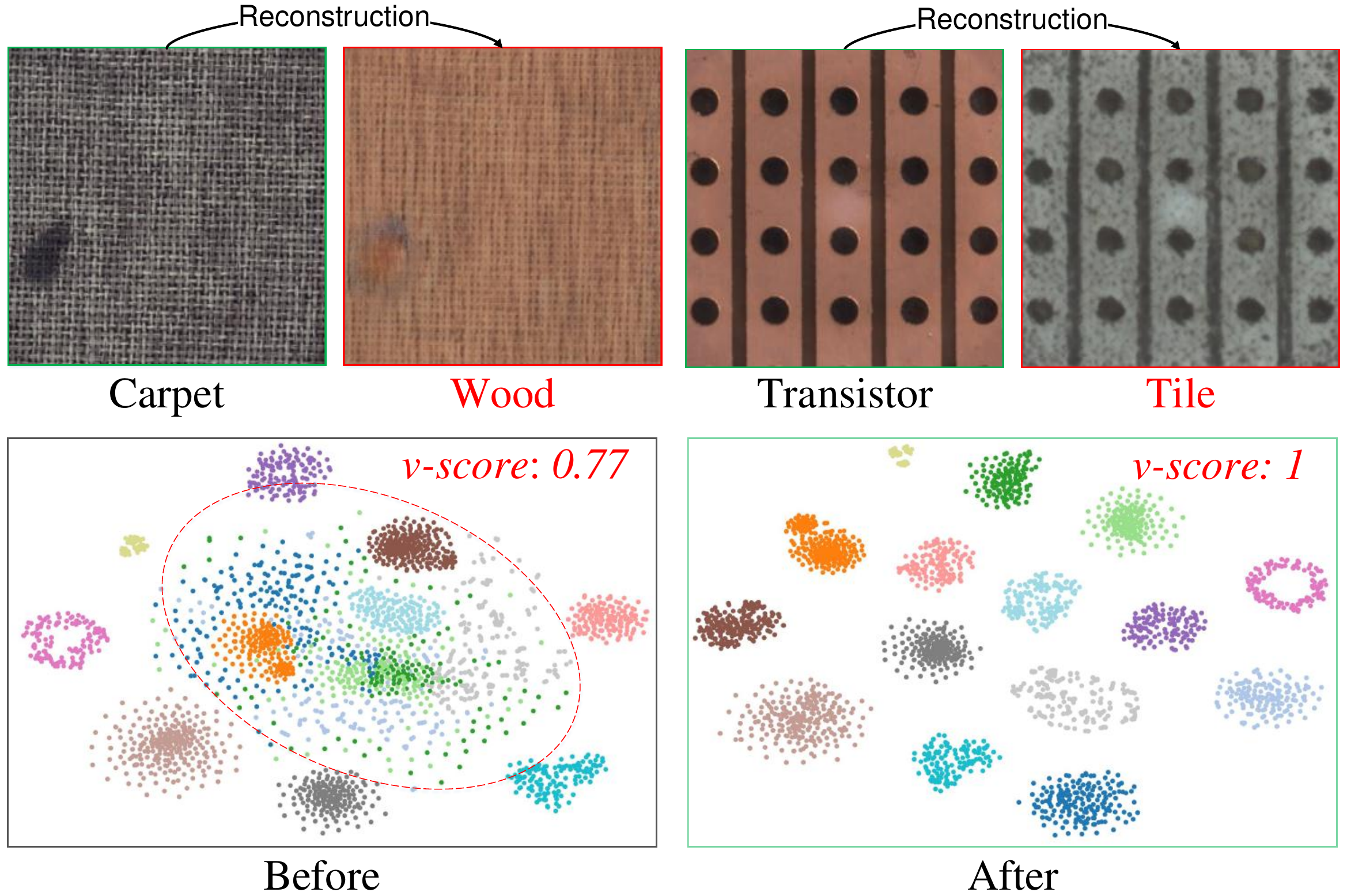} 
        \caption{Reconstruction models trained on mixed data \cite{mvtec_ad} incorrectly reconstruct Carpet \textcolor{orange}{ $\not\rightarrow$ Wood} and Transistor \textcolor{orange}{$\not\rightarrow$ Tile}. In the t-SNE visualizations \cite{van2008visualizing}, multiple classes are initially entangled, but applying our method (CCL) achieves a clear separation, with a V-score of 1.}
        
    \end{subfigure}
    
    \caption{\textbf{(a) CCL} enhances the performance of one-for-one models in multi-class settings, and \textbf{(b) inter-class confusion} arises when reconstruction-based models are trained on multi-class data.}
    \label{fig:challenges}
    \vspace{-1em}
\end{figure}

These methods \cite{rd4ad,liu2023simplenet,rd++,zavrtanik2021draem,zhang2023destseg} typically train separate models for each category, achieving remarkable performance on various datasets \cite{mvtec_ad,visa_ad}. This approach is referred to as the `one-for-one' training scheme. However, it poses several challenges in real-world applications due to difficulties in model management, computational costs, and limited scalability as the number of categories increases. Additionally, these methods require extra information to determine which model should be utilized during inference. To address these limitations, recent efforts have focused on the `one-for-all' setting, where a single model is trained to perform anomaly detection across multiple categories \cite{huang2022registration,uniAD,zhao2023omnial,lee2024crad}. However, these one-for-all models employ larger models and more complex structures but produce suboptimal performance compared to one-for-one models for each class.

A straightforward idea is to extend early one-for-one models to a one-for-all setting while maintaining their performance for each class. However, directly training these one-for-one models on multiple classes often produces heavily degraded results \cite{uniAD,zhao2023omnial}. This raises the question, ``\textit{Why do one-for-one models degrade when trained on multiple classes?}'' To explore this, we conducted empirical experiments primarily using two reconstruction-based methods \cite{rd4ad,zhang2023destseg}, with three training strategies: \textit{Sequential} (training on each class consecutively), \textit{Continual} (training with continual learning \cite{tang2024incremental}), and \textit{Joint} (training with all samples mixed \cite{wang2024comprehensive}), as illustrated in Fig.~\ref{fig:challenges}a. We find that both models perform well for each class under a one-for-one scheme. However, \textit{Sequential} fails to maintain this performance, indicating that the model struggles to retain previously learned knowledge as new classes are introduced, leading to the phenomenon of catastrophic forgetting \cite{liu2024unsupervised,li2022towards}. \textit{Continual} and \textit{Joint} can be interpreted as forms of experience replay \cite{parisi2019continual}. Although both offer improvements compared to \textit{Sequential}, their performance remains significantly below that of the one-for-one approach.

We further conducted a qualitative investigation into the performance degradation of \textit{Joint} by comparing original input images with their reconstructed outputs. This analysis revealed a fundamental issue of \textbf{inter-class confusion}, as illustrated in Fig.~\ref{fig:challenges}b. For example, models incorrectly reconstructed an input image of the `carpet' class as `wood' or misinterpreted `transistor' as `tile'. The model struggled to maintain accurate texture styles, particularly when anomalies exhibited stylistic similarities to other classes (\eg tile). To further understand this, we visualized the feature space using truncated encoder features with t-SNE \cite{van2008visualizing}, revealing that when trained on mixed data, different classes became entangled with a lower V-score. This \textit{inter-class confusion} significantly hinders the model’s ability to reconstruct images accurately and localize anomalous regions.

In this paper, we aim to enhance reconstruction-based models for multi-class anomaly detection. The core idea is to explicitly leverage the category information (\ie the object class: carpet or wood), which is often disregarded by previous methods \cite{uniAD,huang2022registration,zhao2023omnial,gao2024onenip}. To do this, we propose a simple modification, termed local and global Class-aware Contrastive Learning (CCL). We utilize the class information of samples, \eg the 15 object categories in MVTec, as supervised signals to construct positive and negative pairs across different classes for Contrastive Learning (CL). Given a reconstruction-based model comprising an encoder, a bottleneck, and a decoder, the encoder and bottleneck sequentially extract multiscale and compressed features from input images. 
Specifically, \textit{local CL} is applied at the local feature level, for each feature vector extracted from different spatial positions within the multiscale features. We identify nearby spatial positions and search for the most similar feature vectors from other same-class samples, treating them as positive pairs for contrastive learning. This approach encourages the model to capture subtle, class-specific normal patterns. In contrast, \textit{global CL} is applied to the image-level compressed features to align representations within the same class while separating those of different classes, resulting in more compact representations for each class. By integrating class information, both local and global CL encourage the model to capture class-aware and compact representations, effectively mitigating the issue of inter-class confusion. 

Our contributions are as follows:

\begin{itemize} 
\item We extend existing one-for-one reconstruction methods to multiclass anomaly detection by explicitly incorporating raw class information through a simple yet effective modification, mitigating inter-class confusion.
\item  We introduce local and global Class-aware Contrastive Learning (CCL) to effectively enhance class-aware feature representations.
\item Extensive experiments on MVTec \cite{mvtec_ad}, VISA \cite{visa_ad}, BTAD \cite{btad_ad}, Real-IAD \cite{real_ad}, and MANTA~\cite{fan2025manta} datasets demonstrate the effectiveness of our CCL, achieving superior performance compared to advanced methods. Moreover, CCL exhibits strong potential for enhancing one-for-all models and extending effectively to 3D UAD scenarios.
\end{itemize}

\section{Related Work}
\subsection{Visual Anomaly Detection}

\textbf{One-for-one models}. Early methods \cite{xia2022gan,pang2021deep} focused on training a separate model for each class, a strategy known as the one-for-one setting, which specializes in detecting anomalies within a single category. These models can be categorized into three directions: reconstruction-based, synthesis-based, and embedding-based methods. Specifically, reconstruction-based methods \cite{rd4ad,zavrtanik2021draem,liu2023diversity,zhang2023destseg} operate on the assumption that models trained exclusively on normal samples will produce higher reconstruction errors for anomalous regions. Synthesis-based methods \cite{li2021cutpaste,ristea2022self,fan2024patch,zavrtanik2021draem,zhang2023destseg,fan2025grainbrain,fan2023identifying,zhang2023prototypical,zhang2024realnet,lu2023removing} convert AD into a supervised task by training a classifier to distinguish between normal and pseudo-anomalous samples generated through noise injection. Embedding-based methods utilize pretrained models~\cite{su2021bcnet} to extract features from normal samples and model their density using approaches such as memory \cite{patchcore}, Gaussian distribution \cite{defard2021padim}, and normalizing flows \cite{gudovskiy2022cflow}.

\textbf{One-for-all models}. Recent studies \cite{uniAD,guo2025dinomaly} have shifted towards training a unified model capable of handling multiple classes, enabling more scalable and generalizable anomaly detection across diverse categories within a single framework. RegAD \cite{huang2022registration} introduces a registration framework to align input images, highlighting anomalous regions through comparative analysis. Approaches, \eg UniAD \cite{uniAD} and IUF \cite{tang2024incremental}, integrate Transformer architectures~\cite{su2022vitas} into reconstruction-based methods. Meanwhile, OmniAL \cite{zhao2023omnial} and DiAD \cite{he2024diffusion} utilize synthetic data generation for model training, whereas CRAD \cite{lee2024crad} and HGAD \cite{HGAD} employ density distribution methods. However, these one-for-all models produce suboptimal results compared to one-for-one models. 

We analyzed the limitations of previous reconstruction-based models and identified the key challenge in extending them to multi-class scenarios. We thus modified these models by explicitly leveraging class information.

\subsection{Contrastive Learning in AD}

Contrastive learning (CL) \cite{jing2020self,liu2021self} aims to learn task-agnostic feature representations by aligning similar pairs while separating dissimilar pairs in the feature space. This process encourages models to capture meaningful patterns from data without requiring labeled samples, making it widely applicable across various downstream applications \cite{gui2024survey,10904852}. In UAD, CSI \cite{tack2020csi} treats different samples within the same category as negative pairs for novelty detection, whereas Liao \etal \cite{liao2024coft} used synthesized anomaly samples as negative pairs to fine-tune the pretrained models. Recently, ReContrast \cite{guo2024recontrast} incorporates CL to bridge the domain gap. UCAD \cite{liu2024unsupervised} and ReConPatch \cite{hyun2024reconpatch} employ CL to improve structure identification for anomalous regions.  

We introduce class-aware CL by leveraging class information to construct sample pairs within and across different classes, encouraging models to capture normal patterns and enhancing performance in the one-for-all setting.

\section{Methods}

We begin by summarizing recent reconstruction-based models and presenting a generalized one-for-all setting. Next, we propose a simple yet effective improvement strategy that integrates class-aware contrastive learning.

\subsection{Preliminaries}

\begin{figure}[t]
	\centering
	\includegraphics[width=0.44\textwidth]{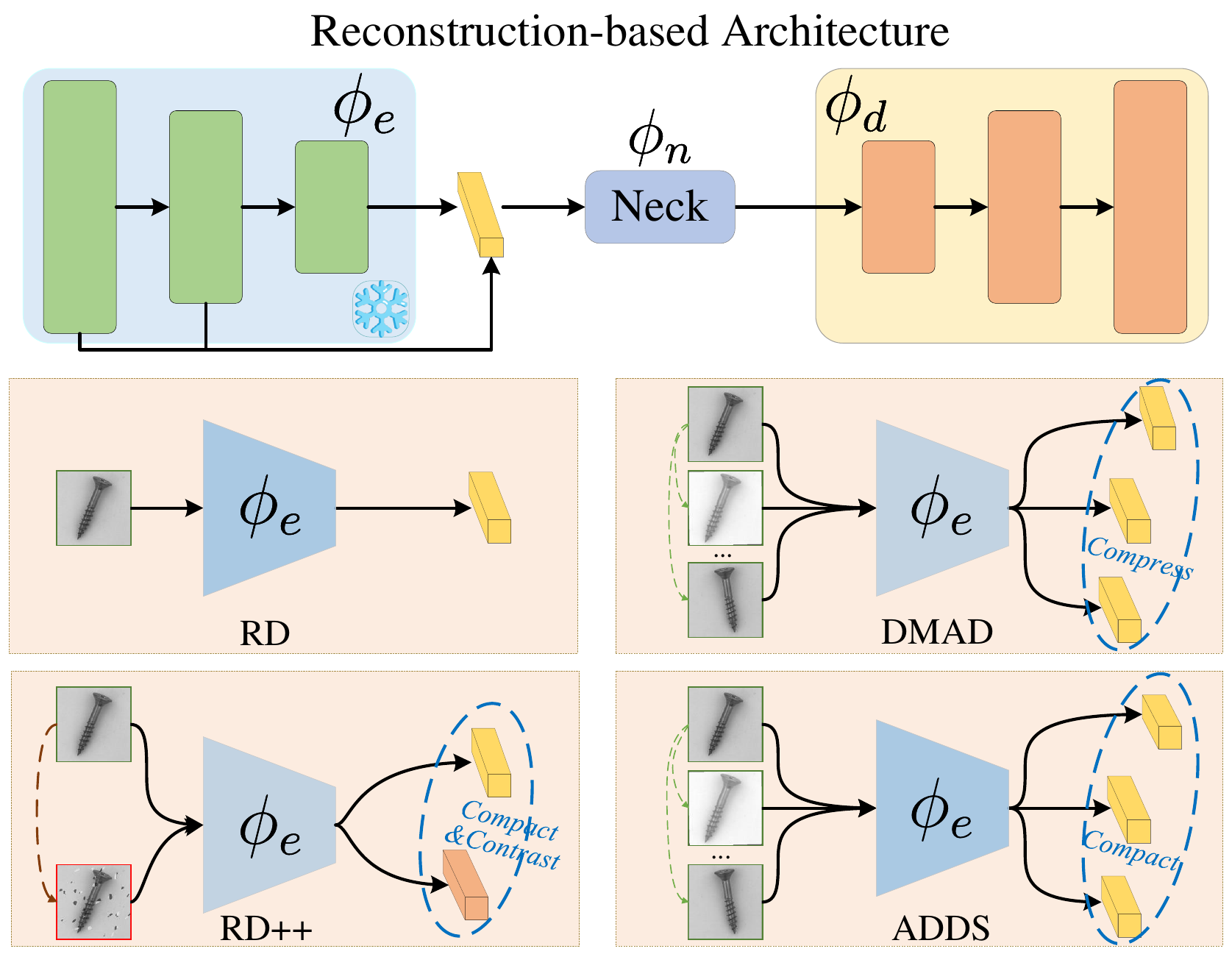}
	\caption{The reconstruction-based model (\ie RD \cite{rd4ad}) has evolved with methods: DMAD \cite{liu2023diversity}, RD++ \cite{rd++}, and ADDS \cite{adds}, which generate \textcolor[RGB]{60, 179, 113}{augmented} or \textcolor[RGB]{132, 60, 20}{pseudo-anomalous} images and regularize their features using \textcolor[RGB]{0, 112, 192}{various constraints}.} 
	\label{fig:reconstruction}
    \vspace{-1em}
\end{figure}

\textbf{Reconstruction-Based Models.} 
The core idea is to train an encoder-decoder model to reconstruct inputs using only normal samples, leading to higher reconstruction errors for anomalous samples during testing.  
A prominent model, RD \cite{rd4ad}, as depicted in Fig.~\ref{fig:reconstruction}, consists of an encoder ($\phi_e$), a trainable one-class neck ($\phi_n$), and a decoder ($\phi_d$). The encoder $\phi_e$ leverages a fixed pretrained model to extract multi-scale features from input images. These features are processed through the neck $\phi_n$, which serves as an information bottleneck to compress them into compact feature representations. The decoder $\phi_d$ then reconstructs results from these compact features, enabling anomaly detection by identifying discrepancies between the input and results.

Given an input image $\mathcal{I}$, multiscale features $\mathbf{f}=\{f_i\}$ (extracted from the $i$-th stage) and compact features $\mathbf{z}$ are obtained sequentially by passing $\mathcal{I}$ through $\phi_e(\mathcal{I})$ and then processing $\phi_n(\mathbf{f})$. Recent studies have extended RD by employing additional augmented samples to achieve more compact feature representations. DMAD \cite{liu2023diversity} and ADDS \cite{adds} generate multiple augmented versions of $\mathcal{I}$, while RD++ \cite{rd++} synthesizes pseudo-anomalous samples. These methods then compress, compact, or contrast these features to establish tighter feature boundaries. However, these enhancements are tailored to the one-for-one setting without considering the variability across different classes.

\begin{figure*}[t]
	\centering
	\includegraphics[width=0.96\textwidth]{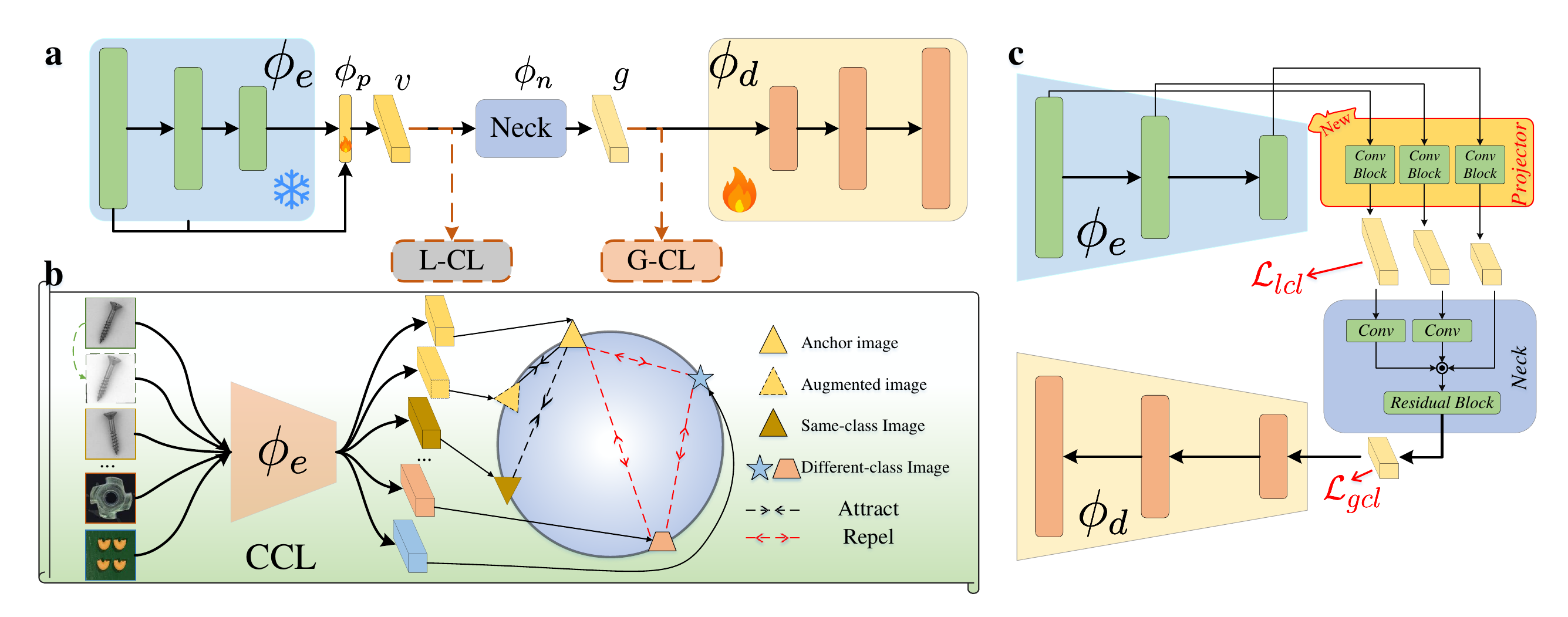}
	\caption{\textbf{a. Overview of CCL.} Considering reconstruction-based models, we employ a projector $\phi_p$ following the encoder $\phi_e$ and apply both local CL and global CL around the neck to learn compact feature representations for each class. \textbf{b. Positive pair selection.} For each anchor image, its augmented version and other samples from the same class are treated as positive pairs. \textbf{c. Detailed structure.} Building upon RD \cite{rd4ad} as the backbone, our CCL incorporates only a projector and two losses (highlighted in red) to effectively adapt the original one-for-one method to one-for-all settings.}

	\label{fig:model}
\end{figure*}

\textbf{One-for-one to One-for-all Objectives.}
We extend the one-for-all scenario to encompass multiple datasets from different domains, represented by a mixed training set $\mathcal{S}_T=\{ \mathcal{I}_t\}_{t=1}^{M}$ and a test set $\mathcal{S}_Q=\{ \mathcal{I}_q \}_{q=1}^{N}$, each spanning a total of $C$ classes. For example, MVTec \cite{mvtec_ad} and Real-IAD \cite{real_ad} consist of 15 and 30 object categories, respectively. When combined, the mixed dataset results in $C = 45$ distinct categories. The training set $\mathcal{S}_T$ contains $M$ normal samples, while $\mathcal{S}_Q$ consists of $N$ samples, which can be either normal or anomalous, with each sample $\mathcal{I}_i$ assigned to a specific class $c_i$. The objective is to train a single model capable of detecting anomalies across all $C$ categories. We aim to enhance the reconstruction-based model $\left< \phi_e, \phi_n, \phi_d \right>$ to capture the feature distribution of normal samples using a mixed training set $\mathcal{S}_T$ across all $C$ classes. During inference, a sample $\mathcal{I}_q$ is classified as anomalous if it produces relatively higher reconstruction errors.

\subsection{Class-aware Contrastive Learning}

We adapt existing reconstruction-based models to the one-for-all setting through three key modifications: a class-aware sampling strategy, local CL, and global CL. We refer to these improvements as local and global Class-aware Contrastive Learning (CCL), as shown in Fig.~\ref{fig:model}.a and c.

\textbf{Class-aware Training Strategy}.
In the one-for-one setting, models are trained using only samples from a single class, while existing one-for-all models \cite{uniAD,zhao2023omnial,lee2024crad,HGAD,gao2024onenip} are typically trained on mixed multiclass data but disregard the original object category information. In this work, we explicitly retain and utilize class information by employing class-aware CL \cite{liu2021self} to achieve more compact and tighter feature representations for each class. Unlike classical CL methods \cite{oord2018representation,chen2020simple}, which treat each instance individually, we consider normal samples from the same class in AD to share a majority of their characteristics, inherently qualifying them as positive pairs \cite{khosla2020supervised,tian2020makes}.

For example, given an input $\mathcal{I}^{c_1}_a$ as an anchor image, we form a tuple $\left<\mathcal{I}^{c_1}_a, \mathcal{I}^{c_1}_{a'}, \mathcal{I}^{c_1}_{b}, \mathcal{I}^{c_2}_c\right>$, where $\mathcal{I}^{c_1}_{a'}$ is an augmented version generated through random data augmentation. Here, $c_1$ and $c_2$ represent different classes. The pairs $\left<\mathcal{I}^{c_1}_a, \mathcal{I}^{c_1}_{a'} \right>$ and $\left<\mathcal{I}^{c_1}_a, \mathcal{I}^{c_1}_{b} \right>$ are treated as positive pairs, while maintaining separation from samples of different classes like $\mathcal{I}^{c_2}_c$. This is achieved by employing a supervised contrastive loss \cite{khosla2020supervised} defined as follows:
\begin{equation}
   \mathcal{L}_{scl}(i) = \frac{-1}{|\mathcal{P}(i)|}\sum_{p\in\mathcal{P}(i)} \log \frac{\exp(\mathbf{f}_i \cdot \mathbf{f}_p/ \tau)}{\sum_{a \in \mathcal{A}(i)} \exp(\mathbf{f}_i \cdot \mathbf{f}_a / \tau)} ,
\end{equation}
where $\cdot$ represents the cosine similarity, $\mathbf{f}_i$ denotes the features extracted from the anchor sample, and $\tau$ is a temperature parameter. $\mathcal{P}(i)$ is the set of positive samples, including augmented features and other samples from the same class, while $\mathcal{A}(i)$ denotes the set of negative samples excluding the anchor $i$ sample (see Fig.~\ref{fig:model}.b).

\textbf{Local Contrastive Learning}.
We apply the class-aware training strategy to multiscale features extracted by the encoder, encouraging the model to capture subtle, class-specific details. To better accommodate multiple classes and domains, we introduce a projector ($\phi_p$, 4 conv blocks) following the encoder to refine and adapt these features.

\begin{figure}[t]
	\centering
	\includegraphics[width=0.48\textwidth]{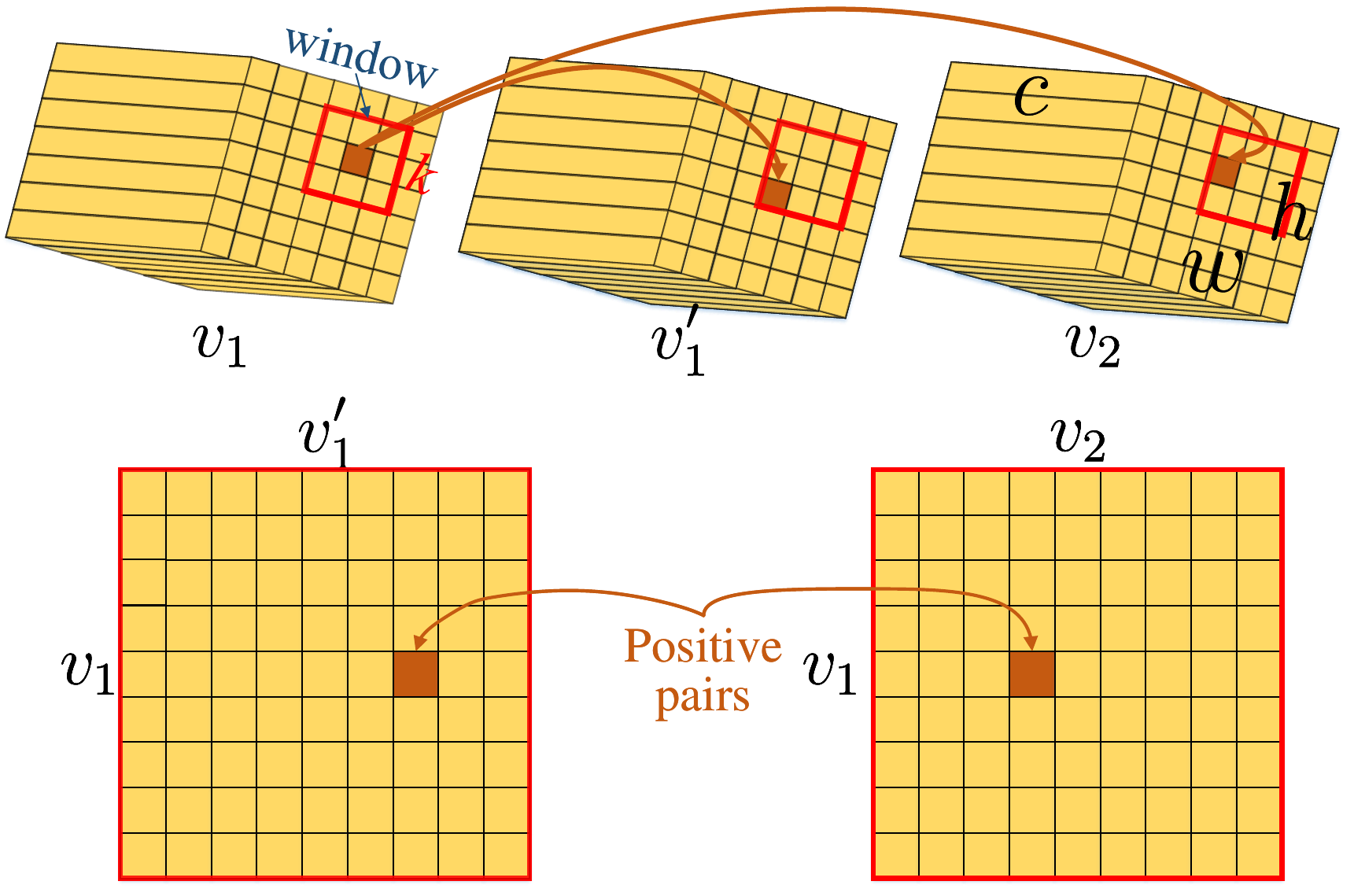}
	\caption{\textbf{Local Contrastive Learning}. For spatial position $(i, j)$ within an anchor feature $v_1$, we compute its similarity with the corresponding features in an augmented feature $v^{'}_1$ and another same-class sample $v_2$, constrained to a window of size $k \times k$. The features within this window are flattened, and the highest similarity value in the resulting similarity matrix is identified. The most similar pairs are treated as positive samples.} 
	\label{fig:lcl}
\end{figure}

Local CL is applied within each individual class. For an anchor image $\mathcal{I}_a$, its augmented version $\mathcal{I}^{'}_a$, and another image $\mathcal{I}_b$ from the same class, features are extracted using the encoder and projector, resulting in representations $v\in\mathbb{R}^{h\times w \times c}$. We measure the similarity between features of positive pairs (\eg $v_a$ and $v^{'}_a$) at each spatial position. The similarity is defined as follows:
\begin{equation}
\mathbf{S}_{(x,y), (m,n)} = \frac{v_{a,(x,y)} \cdot v^{'}_{a,(m,n)}}{\|v_{a,(x,y)}\| \cdot \|v^{'}_{a,(m,n)}\|} ,
\end{equation}
where $(m, n) \in \mathcal{N}_{k}(x,y)$ represents a window of size $k\times k$ centered at position $(x,y)$ within a radius of $\lfloor k/2 \rfloor$, as shown in Fig.~\ref{fig:lcl}. The index of the highest similarity within the window is determined by:
\begin{equation}
\text{index}_{\max}((x,y)) = \arg\max_{(m, n) \in \mathcal{N}_{k}(x,y)} \mathbf{S}_{(x,y), (m,n)},
\end{equation}
where the local features corresponding to $\text{index}_{\max}((x,y))$ are treated as positive pairs. $\mathcal{P}(v(i))$ includes all positive samples, while $\mathcal{A}(v(i))$ denotes the set of negative samples, including all other spatial features. This approach restricts the similarity calculation to a localized window. The positive sample is determined directly at the same spatial position when $k=1$. Conversely, when $k=max(h,w)$, the positive sample match spans all spatial positions. The local CL loss $\mathcal{L}_{lcl}$ is defined as:
\begin{equation}
\mathcal{L}_{lcl} = \sum_{i\in \mathcal{B}}\frac{1}{|hw|}\sum_{(x,y)\in \mathcal{N}_{v}} \mathcal{L}_{scl}(v(i)_{(x,y)}) ,
\end{equation}
where $\mathcal{N}_{v}$ denotes all spatial positions across the feature map, and $\mathcal{B}$ represents a training batch of samples.

Local CL is inspired by embedding-based methods \cite{defard2021padim,gudovskiy2022cflow,wang2021dense} that model the distribution of positional patches. It is based on the observation that samples from the same class often share spatial information across similar views. By applying $\mathcal{L}_{lcl}$, local features from the same class are effectively compacted, enhancing intra-class feature consistency while boosting the discriminative capability against unrelated patches.

\textbf{Global Contrastive Learning}.
We further apply the class-aware training strategy to global features $g \in \mathbb{R}^{c}$ extracted by the neck, promoting separation between different classes. For example, given an anchor feature $g^{c_1}_a$, its augmented version $g^{c_1}_{a^{'}}$, another same-class sample $g^{c_1}_b$, and a different-class sample $g^{c_2}_{c}$, $\mathcal{P}(g(i))$ denotes the set of positive samples, which includes all pairs within the same class. $\mathcal{A}(g(i))$ represents the set of negative samples excluding $g(i)$. The global CL loss $\mathcal{L}_{gcl}$ is defined as:
\begin{equation}
\mathcal{L}_{gcl} = \sum_{i \in \mathcal{B}} \mathcal{L}_{scl}(g(i)) .
\end{equation}

The total training objective of CCL $\mathcal{L}_{total}$ is defined as:
\begin{equation}
\mathcal{L}_{total} = \mathcal{L}_{KD} + \lambda_1\mathcal{L}_{lcl} + \lambda_2\mathcal{L}_{gcl},
\end{equation}
where $\mathcal{L}_{KD}$ represents the loss of the original model (\ie RD \cite{rd4ad}), and $\lambda_1$ and $\lambda_2$ are weights for the respective loss terms. To ensure effective learning and capture meaningful distinctions, each training batch includes samples from different classes, preserving class information to strengthen feature boundary separation.

\subsection{Analysis of One-for-all Challenges}

In the generalized multi-class setting, existing reconstruct ion-based methods \cite{rd4ad,rd++, adds, zhang2023destseg} rely on a fixed pretrained encoder for feature extraction. In the one-for-one setting, data is processed through a one-class bottleneck, enabling the model to effectively learn the distributions of normal samples for a specific class. The decoder can successfully reconstruct the original images due to its flexibility and redundancy. However, extending these models to multiple classes often results in rapid performance degradation, primarily due to significant domain gaps between the pretrained model and features from diverse domains and classes. To mitigate this issue, we introduce a class-aware training strategy that incorporates mixed datasets and leverages local contrastive learning with a projector to bridge these feature gaps.

\textbf{Inter-Class Confusion}.
When reconstruction results exhibit confusion across different classes, we attribute this to the neck's inability to effectively retain distinct feature representations for each class. Ideally, multiple classes should remain clearly distinguishable, while features within the same class should demonstrate lower variability compared to those from different domains. To this end, we introduce global contrastive learning to impose explicit feature constraints via a class-aware training strategy. This approach tightens class boundaries and prevents the decoder from incorrectly interpreting features during reconstruction, thereby reducing inter-class confusion.

\begin{table*}[t]
\caption{\textbf{Comparison of various one-for-all models}. \textit{Domain-in-all} refers to combining all classes within each dataset, while \textit{All-in-all} signifies combining all classes across MVTec AD, Visa, BTAD, and Real-IAD datasets. Results are reported as I-AUROC / P-AUROC / PRO ($\%$). $C$ is the number of classes. * indicates that the standard deviations across five random seeds are smaller than 0.1.}
 \centering
\resizebox{0.98\textwidth}{!}{
    \begin{tabular}{lcccccc}
    \toprule
    \toprule
    \multirow{2}[0]{*}{Model('year)} & \multicolumn{5}{c}{\textit{Domain-in-all}}     & \multirow{2}[0]{*}{\textit{All-in-all ($C$=60)}} \\
    \cmidrule(lr){2-6}
          & MVTec AD ($C$=15) & Visa ($C$=12)  & BTAD ($C$=3)  & Real-IAD ($C$=30) & MANTA ($C$=38) &  \\
    \cmidrule(lr){1-7}
    UniAD'22 (NeurIPS \cite{uniAD}) & 97.5 / 97.0 / 90.7 & 88.8 / 98.3 / 85.5 & 91.9 / 95.5 / 75.1 & 82.9 / 97.6 / 86.4 & 79.3 / 89.3 / 76.8 & 86.5 / 96.3 / 86.9 \\
    CRAD'24 (ECCV \cite{lee2024crad})  & {99.3 / 97.8 / 91.7} & 94.7 / 98.1 / 84.0 & 92.5 / 97.1 / 72.5 & \textbf{88.5} / 97.7 / 85.8 & \textbf{88.9} / 93.0 / 81.5 & 89.5 / 97.7 / 86.5 \\
    OneNIP'24 (ECCV \cite{gao2024onenip}) & 97.9 / 97.9 / 93.4 & 92.5 / {98.7} / 84.8 & 92.6 / 97.4 / 76.5 & 83.9 / 96.7 / 85.0 & 79.5 / 90.2 / 78.7 & 86.1 / 96.1 / 83.5 \\
    DiAD'24 (AAAI \cite{he2024diffusion})  & 97.2 / 96.8 / 90.7 & 86.8 / 96.0 / 75.2 & 92.6 / 97.3 / 76.5 & 75.6 / 88.0 / 58.1 & 74.7 / 86.8 / 74.2 & 85.2 / 95.3 / 82.9 \\
    {RLR'24} (ECCV \cite{he2024learning})  & 98.3 / 97.8 / 92.1 & 93.8 / \textbf{99.0} / \textbf{93.8} & 94.3 / 96.9 / 75.1 & 86.4 / 98.4 / 88.0 & 83.1 / 92.1 / 80.8 & 88.4 / 97.6 / 88.6 \\
    MambaAD'24 (NeurIPS \cite{he2024mambaad}) & 98.6 / 97.7 / 93.1 & 94.3 / 98.5 / 91.0 & 95.0 / 97.8 / 78.9 & 86.3 / 98.5 / 90.5 & 80.0 / 91.6 / 76.9 & 89.2 / 97.7 / 90.3  \\
    \cmidrule(lr){1-7}
    CCL (+RD, our) & \textbf{99.3} / \textbf{98.2} / \textbf{93.6} & \textbf{95.9} / 98.5 / 92.6 & \textbf{95.7} / \textbf{98.0} / \textbf{80.2} & 87.6 / \textbf{98.5} / \textbf{91.4} & 84.3 / \textbf{93.7} / \textbf{81.9} & \textbf{90.6}*/ \textbf{97.9}/*\textbf{90.6}*  \\
    
    \bottomrule
    \bottomrule
    \end{tabular}%
}
\label{tab:sota}
\end{table*}

\section{Experiments}

\subsection{Dataset and Implementation Details}

\textbf{Datasets}. We conducted experiments on five widely-used anomaly detection datasets: \textbf{MVTec AD} \cite{mvtec_ad}, consisting of 3,629 training and 1,725 test high-resolution images across 15 classes of industrial objects and textures; \textbf{VISA} \cite{visa_ad}, including 8,659 training samples and 2,162 test samples from 12 diverse objects; \textbf{BTAD} \cite{btad_ad}, comprising three types of industrial products with 1,799 training samples and 741 test samples; \textbf{Real-IAD} \cite{real_ad}, covering 30 objects with 36,465 training and 114,585 test multi-view images; and \textbf{MANTA} \cite{fan2025manta} having 137.3K images across 38 object categories.

To evaluate model performance comprehensively in a one-for-all setting, we define two configurations: \textit{\textbf{domain-in-all}}, where all categories within a single dataset are mixed for evaluation, and \textit{\textbf{all-in-all}}, where data from multiple datasets are combined for evaluation.

\noindent \textbf{Implementation Details.} Our backbone was mainly based on RD \cite{rd4ad}. The batch size was set to 16, $\tau$ was set to 0.1, and the learning rates were 0.001 and 0.005 for the projector and other parts with the Adam optimizer. We standardized the input resolution to $256 \times 256$ across all datasets and applied data augmentations, including resizing, flipping, color jittering, and rotation to enhance model robustness. Models were trained with early stopping based on validation loss. During inference, we followed the RD approach by calculating cosine similarity between multi-scale features and applying Gaussian smoothing with $\sigma=4$ for the results.

\textbf{Evaluation Metrics.} We used three metrics \cite{he2024mambaad,he2024diffusion} to evaluate model performance: Image-level AUROC (I-AUROC) and Pixel-level AUROC (P-AUROC), which measure anomaly detection and localization on a per-image basis; and P-AUPRO (PRO), which quantifies the precision-recall trade-off for pixel-level detection. All metrics range from 0 to 1, with higher values indicating better results.

\subsection{Comparison and Results}

We conducted comprehensive experiments across four datasets to compare our model with several state-of-the-art one-for-all models, including UniAD \cite{uniAD}, CRAD \cite{lee2024crad}, OneNIP \cite{gao2024onenip}, DiAD \cite{he2024diffusion}, RLR \cite{he2024learning} and MambaAD \cite{he2024mambaad}. We tested both \textit{domain-in-all} and \textit{all-in-all} settings, as shown in Table~\ref{tab:sota}.
We observed that, under the domain-in-all setting, our model achieved the best performance across nearly all five datasets in all three metrics.
When mixing four datasets (MVTec, Visa, BTAD and Real-IAD) in the \textit{all-in-all} setting, our model continued to outperform, with I-AUROC of 90.6$\%$, P-AUROC of 97.8$\%$ and PRO of 90.5$\%$, demonstrating its robustness to domain shifts and the increase in category diversity compared to other approaches. 

We further evaluated the effectiveness of our framework on the COCOAD dataset \cite{zhang2024learning}, which includes more class categories and provides comprehensive metrics (following the evaluation protocol of MambaAD \cite{he2024mambaad}), as shown in Table~\ref{tab:coco}. Compared to other one-for-all methods \cite{uniAD,he2024mambaad}, our framework achieved the best overall result, demonstrating substantial gains over the original RD \cite{rd4ad}. Notably, our CCL maintains strong performance as the number of classes increases, ranging from BTAD ($C$=3), Visa ($C$=12), MVTec AD ($C$=15), Real-IAD ($C$=30), MANTA ($C$=38), All-in-all ($C$=60), to COCOAD ($C$=81).

We attribute these improvements to our local and global contrastive learning design, which not only enhances stability in I-AUROC scores but also consistently improves P-AUROC and PRO metrics. Unlike models such as UniAD, OneNIP, and MambaAD, which rely on complex model architectures, or DiAD, which employs heavy data augmentation strategies, our model extends one-for-one reconstruction-based models with a simple convolutional structure. This design makes our model particularly suitable for practical deployment in industrial environments.

\begin{table}[t]
\caption{\textbf{Performance (\%) vs. more metrics and class numbers}. We evaluated our model on the COCOAD dataset \cite{zhang2024learning} ( $C$=81).}
 \centering
\resizebox{0.48\textwidth}{!}{
    \begin{tabular}{c|cccc}
    \toprule \toprule
      Model & I- / P- AUROC  & I- / P- mAP  & I- / P- mF1-max & PRO \\
     \cmidrule(lr){1-5}
    UniAD \cite{uniAD} & 55.2 / 64.6 & 49.3 / 12.8 & 61.7 / 19.0 & 34.3 \\
    MambaAD \cite{he2024mambaad} & 63.9 / 69.3 & 56.2 / 16.9 & 63.2 / 22.2 & 40.5 \\
    RD \cite{rd4ad} & 57.6 / 66.5 & 49.9 / 13.9 & 62.0 / 20.0 & 39.8 \\
     \cmidrule(lr){1-5}
    CCL (with RD) & \textbf{65.2} / \textbf{71.5} &\textbf{57.8} / \textbf{18.6} & \textbf{64.0} / \textbf{24.4} & \textbf{43.5} \\
      \bottomrule \bottomrule
    \end{tabular}%
}
\label{tab:coco}
\end{table}

\subsection{Ablation Studies}

We conducted experiments to evaluate each design component of our CCL on three datasets: MVTec, VISA, and BTAD, under the \textbf{all-in-all} setting, covering a total of 30 categories. The default models were built on RD, utilizing four convolutional blocks (4-conv-blocks) as the projector $\phi_p$, multiscale features including the last three stages ($\mathbf{f}=\{f_1, f_2, f_3\}$), and the window size $k$ was set to match the full size of the feature map.

\begin{table}[t]
\caption{\textbf{Ablation study on individual components and pseudo-classes}. Results are reported as I-AUROC/P-AUROC/PRO (\%) under the \textit{all-in-all} setting across three datasets: MVTec, VISA, and BTAD. $K_C$ represents the number of pseudo-classes.}
 \centering
\resizebox{0.46\textwidth}{!}{
    \begin{tabular}{cccc}
    \toprule \toprule
     {Baseline}  & w. L-CL & w. G-CL  &  CCL (Raw Class)   \\
     91.0/95.9/88.6 & +5.5/+2.1/+2.4 & +4.2/+1.6/+2.0 & +5.7/+2.2/+2.8 \\
     \cmidrule(lr){1-4}
    \multirow{2}[0]{*}{Pseudo-Classes} & $K_C$=15  & $K_C$=30  & $K_C$=60  \\
    & +5.1/+1.9/+2.3 & +5.7/+2.2/+2.8 & +5.0/+2.0/+2.2 \\
      \bottomrule \bottomrule
    \end{tabular}%
}
\vspace{-1em}
\label{tab:abl-label}
\end{table}

\textbf{Each Component and Pseudo-Classes.} 
We first validated the effectiveness of each component in our CCL and examined the impact of pseudo-class labels, as shown in Table \ref{tab:abl-label}. Specifically, we utilized a ResNet-18 pretrained on ImageNet \cite{he2016deep} to extract features from the final layer, and then applied the $K$-means clustering algorithm (with $K_C$ centers). The index of each cluster was assigned as a pseudo-class label to the corresponding sample. Building upon the baseline (an RD model trained on multi-class data), applying either Local CL (w. L-CL) or Global CL (w. G-CL) yielded substantial improvements. Combining both components resulted in the highest performance, achieving gains of +5.7, +2.2, and +2.8\% in I-AUROC, P-AUROC, and PRO, respectively.

When using pseudo labels generated with different numbers of cluster centers ($K_C$ set to 15, 30, or 60), we observed that the best result was achieved at $K_C = 30$. The model performance remained relatively consistent across different choices of $K_C$ compared to the scenario using raw class labels. 
Although our approach utilizes raw class information, which is typically disregarded by existing approaches, these experimental results demonstrate that even using pseudo-class labels derived through clustering can achieve comparable or superior performance, and we tested several traditional clustering methods that showed similar results. This suggests that class information can serve as a valuable ``free lunch'' in the one-for-all setting.

\textbf{Reconstruction-based Models}. We further validated our CCL approach by incorporating it into more one-for-one reconstruction-based models: DR\textit{A}EM \cite{zavrtanik2021draem}, DeSTSeg \cite{zhang2023destseg}, and DMAD \cite{liu2023diversity}, comparing against the \textit{Joint} training strategy.  Additionally, we explored integrating CCL with existing one-for-all models: ViTAD \cite{vitad} and MambaAD \cite{he2024mambaad}, as well as a 3D one-for-one model: CRD \cite{liu2024multimodal} on the MVTec 3D \cite{mvtec3D} that showed similar performance with advanced methods (\eg AST~\cite{rudolph2023asymmetric} and M3DM~\cite{wang2023multimodal}). As shown in Table \ref{tab:abl-reconstruction}, for one-for-one models, our CCL approach consistently outperformed its counterparts using the \textit{Joint} strategy, achieving average improvements of over 5\% in I-AUROC, 5\% in P-AUROC, and 7\% in PRO metrics. For existing one-for-all models, CCL effectively enhanced their performance with minimal computational overhead. Notably, we observed stable improvements on the CRD model evaluated on the MVTec 3D dataset, highlighting the strong generalization potential of our CCL approach.

However, compared to the original one-for-one results, \eg DR\textit{A}EM, using CCL under the one-for-all setting still yields relatively lower performance. We attribute this decrease to these models' reliance on additional discriminators for anomaly detection, where their encoder-decoder architectures primarily focus on feature learning rather than explicitly modeling the distribution of normal samples.

\begin{table}[t]
\caption{\textbf{Results of different reconstruction-based models.} Performance is reported as I-AUROC/P-AUROC/PRO (\%) for one-for-one and one-for-all models under the \textit{all-in-all} setting across MVTec, VISA, and BTAD. CRD is evaluated on MVTec 3D \cite{mvtec3D}. FLOPs indicate the change in computational overhead.}
 \centering
\resizebox{0.49\textwidth}{!}{
    \begin{tabular}{lccc}
    \toprule\toprule
    One-for-one Models & \textit{Joint}  & CCL (Our)   & FLOPs (G) \\
    \cmidrule(lr){1-4}
    DR\textit{A}EM'21  \cite{zavrtanik2021draem} & 81.4/82.1/63.9 & 88.1/89.7/78.4 &  198.2 $\rightarrow$ 200.5  \\
    DMAD'23 \cite{liu2023diversity}  & 82.9/95.3/82.3 & 94.9/97.7/89.9 & 36.4 $\rightarrow$ 38.4   \\
   RD'22 \cite{rd4ad} & 91.0/95.9/88.6 & 96.7/98.1/91.4 & 28.4 $\rightarrow$ 30.7  \\
    DeSTSeg'23 \cite{zhang2023destseg}& 90.4/84.4/64.6 & 92.6/89.4/74.6 & 30.7 $\rightarrow$ 32.1 \\
    \midrule \midrule
   One-for-all Models & Original  & CCL (Our) &  FLOPs (G) \\
   \cmidrule(lr){1-4}
     ViTAD'23 \cite{vitad}& 93.6/97.6/86.0 & 95.2/98.1/87.1 & 9.7 $\rightarrow$ 10.8 \\
     MambaAD'24 \cite{zhang2023destseg}& 93.6/97.2/89.8 & 95.3/97.9/90.8 & 8.3 $\rightarrow$ 9.4 \\
    \midrule \midrule
     3D Models & \textit{Joint}  & CCL (Our) &  FLOPs (G) \\
     \cmidrule(lr){1-4}
    {CRD}'24 (\cite{liu2024multimodal}) & 88.5/99.1/96.9 & 91.6/99.2/97.1 &  147.1 $\rightarrow$ 149.6 \\
    \bottomrule    
    \bottomrule 
    \end{tabular}
    }
\vspace{-1em}
\label{tab:abl-reconstruction}
\end{table}

\begin{table}[b]
\vspace{-1em}
\caption{\textbf{Ablation study on multi-scale features $\mathbf{f}$, window size $k$, and projector $\phi_p$.} Here, $\max(h,w)$ represents the maximum feature size. Results are reported as I-AUROC / P-AUROC / PRO on MVTec, VISA, and BTAD under the \textit{all-in-all} setting.}
 \centering
\resizebox{0.46\textwidth}{!}{
    \begin{tabular}{ccc|c}
    \toprule \toprule
    $\mathbf{f}$ & $k$ & $\phi_p$ & Performance (\%) \\
    \cmidrule(lr){1-4}
    $-$ & $-$ & $-$ & 91.0 / 95.9 / 88.6 \\
    $f_3$ & $\max(h,w)$ & 4-CBlocks & +2.9 / +1.5 / +1.6 \\
    $\{f_2,f_3\}$ & $\max(h,w)$ & 4-CBlocks    & +3.7 / +1.7 / +2.0 \\
    $\{f_1,f_2,f_3\}$ & $\max(h,w)$ & 2-CBlocks    & +4.7 / +1.7 / +2.2 \\
     $\{f_1,f_2,f_3\}$ & $\max(h,w)$ & 2-MLP    & -15.3 / -8.9 / -23.8 \\
    $\{f_1,f_2,f_3\}$ & $\max(h,w)$ & 4-CBlocks    & +5.1 / +1.9 / +2.2 \\
    $\{f_1,f_2,f_3\}$ & 3  & 4-CBlocks   & +5.0 / +1.7 / +1.7 \\
    $\{f_1,f_2,f_3\}$& 1  & 4-CBlocks   & +5.5 / +2.1 / +2.4 \\
      \bottomrule \bottomrule
    \end{tabular}%
}
\label{tab:abl-localCL}
\end{table}

\textbf{Local CL}. 
We evaluated the impact of multiscale features $\mathbf{f}$ extracted from different encoder stages (using $f_3$ alone, combining $\{f_2,f_3\}$, and incorporating all scales $\{f_1,f_2,f_3\}$), varying local window sizes $k$ (set to $\max{(h,w)}$, 3 and 1), and different projector $\phi_p$ settings (including a 2-layer MLP (2-MLP) \cite{liu2023simplenet}, two conv blocks (2-CBlocks) \cite{rd++}, and 4-CBlocks). All evaluations were conducted without global CL, as shown in Table \ref{tab:abl-localCL}.

We observed that incorporating more scales of feature maps generally produced better performance, as multiscale features offer richer spatial information. When using different $k$ values, only minor performance fluctuations were observed, with the best results achieved when $k$=1. We attribute this fluctuation to high-level feature maps, such as $f_3$ with a resolution of $8\times8$, where each location corresponds to a field-of-view of $32\times32$ pixels in the input image. Nearby positions in the same-class samples often exhibit similar visual characteristics, making $k$=1, which directly compares features at the same spatial location, suitable for positive pairs. In practice, using smaller values of $k$ can significantly reduce the complexity of searching for positive pairs. When employing different projectors $\phi_p$, using 4-CBlocks yielded the best performance, achieving an I-AUROC of 96.7\%, P-AUROC of 98.1\%, and PRO of 91.4\%. It can be attributed to the superior ability of convolutional layers to fine-tune spatial features.

\begin{table}[t]
\caption{\textbf{Ablation study of global CL}. Results are reported as I-AUROC / P-AUROC / PRO (\%) in the \textit{all-in-all} setting across MVTec, VISA, and BTAD.}

 \centering
\resizebox{0.46\textwidth}{!}{
    \begin{tabular}{ccc|c}
    \toprule \toprule
    \multicolumn{1}{c}{Triplet \cite{ge2018deep}} & \multicolumn{1}{c}{N-pairs \cite{npairs}} & infoNCE \cite{oord2018representation}    & $\mathcal{L}_{gcl}$ \\
    \cmidrule(lr){1-4}
    91.0/96.7/88.1 & 91.9/96.1/89.2 & 93.9/96.3/89.7 & \textbf{95.2}/\textbf{97.5}/\textbf{90.6} \\
    \bottomrule \bottomrule
    \end{tabular}%
}
\vspace{-1em}
\label{tab:abl-gcl}
\end{table}

\textbf{Global CL}.
The global CL is introduced to encourage features extracted by the neck to maintain compact and distinct normal patterns for each class. We evaluated different forms: including the classic triplet loss \cite{ge2018deep}, N-pairs loss \cite{npairs}, InfoNCE \cite{oord2018representation}, and our proposed Global CL $\mathcal{L}_{gcl}$. The evaluations were performed using models based on RD without integrating local CL. As shown in Table \ref{tab:abl-gcl}, our $\mathcal{L}_{gcl}$ and InfoNCE demonstrated consistent performance improvements of approximately 2\% in I-AUROC compared to the triplet and N-pairs losses. This is attributed to their richer design of positive and negative sample pairs, leading to more effective feature representation learning. Unlike InfoNCE, which considers only each sample and its augmented version as positive pairs, our $\mathcal{L}_{gcl}$ treats all samples within the same class as positive pairs, better aligning with the requirements of anomaly detection tasks.

\textbf{Hyperparameter $\lambda_1$:$\lambda_2$ and $\tau$ in CCL}.
We evaluated the effect of varying the loss-weight ratios ($\lambda_1$:$\lambda_2$), and the temperature parameter $\tau$ (with $\lambda_1$:$\lambda_2$ fixed at 1:1) for both local CL and global CL. As shown in Table \ref{tab:abl-params}, the optimal result was achieved with a 1:1 ratio, suggesting that local and global CL complement each other by compacting spatial and global features. We also observed that the best results were obtained with $\tau$ set to 0.1. 
Variations in parameters within reasonable ranges resulted in minor fluctuations, remaining within 0.5\% across all metrics.

\begin{figure}[t]
	\centering
	\includegraphics[width=.46\textwidth]{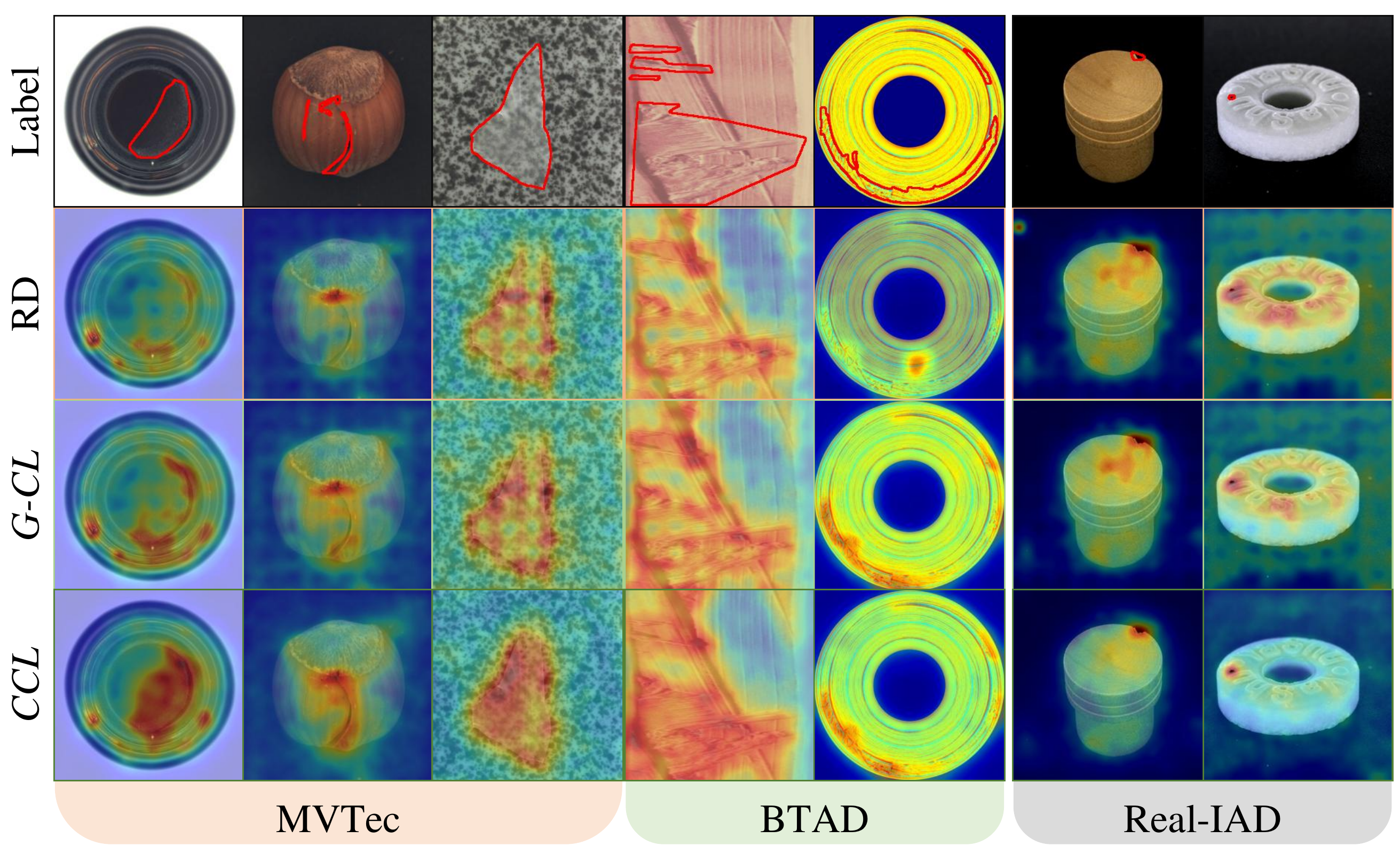}
	\caption{\textbf{Visualization of CCL models on MVTec, BTAD, and Real-IAD datasets.} Label: Test images with anomalous regions highlighted in red contours. RD, Global-CL, and CCL represent the original RD \cite{rd4ad}, our global CL applied to compressed features, and our full CCL, respectively. } 
	\label{fig:vis}    
\end{figure}

\textbf{Visualizations}. We qualitatively visualized our CCL, as illustrated in Fig.~\ref{fig:vis}. Compared to the original RD \cite{rd4ad}, incorporating only global CL effectively improves anomaly detection in large anomalous regions. Building on this, the full CCL further enhances performance, particularly in accurately detecting anomaly boundaries. For example, in MVTec \cite{mvtec_ad}, CCL improved the detection accuracy of cracks in hazelnuts, demonstrating its effectiveness in precise anomaly localization. This improvement is attributed to the local CL capturing compact spatial features. These results qualitatively demonstrate our model's effectiveness in transitioning from a one-for-one to a one-for-all setting.

\begin{table}[t]
\caption{\textbf{Ablation study of the ratios of $\lambda_1$:$\lambda_2$ and $\tau$}. Results are reported as I-AUROC/P-AUROC/PRO (\%) under the \textit{all-in-all} setting across MVTec, VISA, and BTAD.}
 \centering
\resizebox{0.48\textwidth}{!}{
    \begin{tabular}{ccccc}
    \toprule \toprule
    \multirow{2}[0]{*}{\makecell{$\lambda_1:\lambda_2$ \\ ($\tau=0.1$)}} & 0.2   & 0.5   & 1     & 2 \\
    \cmidrule(lr){2-5} 
          & 96.7/98.0/91.2 & 96.6/98.0/91.0 & \textbf{96.7}/\textbf{98.1}/\textbf{91.4} & 96.5/97.9/91.0 \\
    \midrule
    \midrule 
    \multirow{2}[0]{*}{\makecell{$\tau$ \\ ($\lambda_1$:$\lambda_2=1$)}} & 0.05  & 0.07  & 0.1   & 0.2 \\
     \cmidrule(lr){2-5} 
          & 96.3/97.9/90.8 & 96.5/98.0/91.2 & \textbf{96.7}/\textbf{98.1}/\textbf{91.4} & 96.6/98.0/91.2 \\
    \bottomrule   \bottomrule    
     \end{tabular}%
}
\vspace{-1em}
\label{tab:abl-params}
\end{table}

\section{Conclusion}

We presented a simple yet effective approach to enhance one-for-one reconstruction-based models for multi-class settings by incorporating class information through class-aware contrastive learning. Experiments on four datasets validated the effectiveness of our enhancements. 

\textbf{Limitations}. Our proposed CCL has demonstrated effectiveness not only in both one-for-one and one-for-all reconstruction-based methods, but also in 3D one-for-one models. We also explored the generalizability to unseen data, but the result is not satisfactory. We further evaluated the applicability of CCL to synthesis-based (\eg SimpleNet) and embedding-based (\eg CFlow) approaches, but only local CL resulted in marginal gains. We believe this is because these approaches inherently include classifiers designed to distinguish positive from negative (pseudo) samples. Therefore, further improvements would require the joint optimization of these classifiers alongside CCL, representing a promising direction for future research.

{
    \small
    \bibliographystyle{ieeenat_fullname}
    \bibliography{main}

\begin{thebibliography}{67}
\providecommand{\natexlab}[1]{#1}
\providecommand{\url}[1]{\texttt{#1}}
\expandafter\ifx\csname urlstyle\endcsname\relax
  \providecommand{\doi}[1]{doi: #1}\else
  \providecommand{\doi}{doi: \begingroup \urlstyle{rm}\Url}\fi

\bibitem[Bergmann et~al.(2019)Bergmann, Fauser, Sattlegger, and Steger]{mvtec_ad}
Paul Bergmann, Michael Fauser, David Sattlegger, and Carsten Steger.
\newblock {MVTec AD--A comprehensive real-world dataset for unsupervised anomaly detection}.
\newblock In \emph{CVPR}, pages 9592--9600, 2019.

\bibitem[Bergmann et~al.(2021)Bergmann, Jin, Sattlegger, and Steger]{mvtec3D}
Paul Bergmann, Xin Jin, David Sattlegger, and Carsten Steger.
\newblock The mvtec 3d-ad dataset for unsupervised 3d anomaly detection and localization.
\newblock \emph{preprint arXiv:2112.09045}, 2021.

\bibitem[Cao et~al.(2023)Cao, Zhu, and Pang]{adds}
Tri Cao, Jiawen Zhu, and Guansong Pang.
\newblock Anomaly detection under distribution shift.
\newblock In \emph{ICCV}, pages 6511--6523, 2023.

\bibitem[Chen et~al.(2020)Chen, Kornblith, Norouzi, and Hinton]{chen2020simple}
Ting Chen, Simon Kornblith, Mohammad Norouzi, and Geoffrey Hinton.
\newblock A simple framework for contrastive learning of visual representations.
\newblock In \emph{ICML}, pages 1597--1607. PMLR, 2020.

\bibitem[Cheng et~al.(2025)Cheng, Li, Di, et~al.]{cheng2025open}
Wenxiao Cheng, Xue Li, Donglin Di, et~al.
\newblock An open-set semi-supervised contrastive learning for bearing fault diagnosis.
\newblock \emph{IEEE TIM}, 2025.

\bibitem[Defard et~al.(2021)Defard, Setkov, Loesch, and Audigier]{defard2021padim}
Thomas Defard, Aleksandr Setkov, Angelique Loesch, and Romaric Audigier.
\newblock Padim: a patch distribution modeling framework for anomaly detection and localization.
\newblock In \emph{ICPR}, pages 475--489. Springer, 2021.

\bibitem[Deng and Li(2022)]{rd4ad}
Hanqiu Deng and Xingyu Li.
\newblock Anomaly detection via reverse distillation from one-class embedding.
\newblock In \emph{CVPR}, pages 9737--9746, 2022.

\bibitem[Fan et~al.(2022)Fan, Ding, Fan, et~al.]{fan2022grainspace}
Lei Fan, Yiwen Ding, Dongdong Fan, et~al.
\newblock Grainspace: A large-scale dataset for fine-grained and domain-adaptive recognition of cereal grains.
\newblock In \emph{CVPR}, pages 21116--21125, 2022.

\bibitem[Fan et~al.(2023)Fan, Ding, Fan, et~al.]{fan2023identifying}
Lei Fan, Yiwen Ding, Dongdong Fan, et~al.
\newblock Identifying the defective: Detecting damaged grains for cereal appearance inspection.
\newblock In \emph{ECAI}, pages 660--667. IOS Press, 2023.

\bibitem[Fan et~al.(2024)Fan, Ding, Pagnucco, and Song]{fan2024patch}
Lei Fan, Yiwen Ding, Maurice Pagnucco, and Yang Song.
\newblock Patch-wise augmentation for anomaly detection and localization.
\newblock In \emph{ICASSP}, pages 5425--5429. IEEE, 2024.

\bibitem[Fan et~al.(2025{\natexlab{a}})Fan, Fan, Ding, Wu, Di, Pagnucco, and Song]{fan2025grainbrain}
Lei Fan, Dongdong Fan, Yiwen Ding, Yong Wu, Donglin Di, Maurice Pagnucco, and Yang Song.
\newblock Grainbrain: Multiview identification and stratification of defective grain kernels.
\newblock \emph{IEEE TII}, 2025{\natexlab{a}}.

\bibitem[Fan et~al.(2025{\natexlab{b}})Fan, Fan, Hu, et~al.]{fan2025manta}
Lei Fan, Dongdong Fan, Zhiguang Hu, et~al.
\newblock Manta: A large-scale multi-view and visual-text anomaly detection dataset for tiny objects.
\newblock In \emph{CVPR}, pages 25518--25527, 2025{\natexlab{b}}.

\bibitem[Gao(2024)]{gao2024onenip}
Bin-Bin Gao.
\newblock Learning to detect multi-class anomalies with just one normal image prompt.
\newblock In \emph{ECCV}, 2024.

\bibitem[Ge(2018)]{ge2018deep}
Weifeng Ge.
\newblock Deep metric learning with hierarchical triplet loss.
\newblock In \emph{ECCV}, pages 269--285, 2018.

\bibitem[Gudovskiy et~al.(2022)Gudovskiy, Ishizaka, and Kozuka]{gudovskiy2022cflow}
Denis Gudovskiy, Shun Ishizaka, and Kazuki Kozuka.
\newblock Cflow-ad: Real-time unsupervised anomaly detection with localization via conditional normalizing flows.
\newblock In \emph{WACV}, pages 98--107, 2022.

\bibitem[Gui et~al.(2024)Gui, Chen, Zhang, et~al.]{gui2024survey}
Jie Gui, Tuo Chen, Jing Zhang, et~al.
\newblock A survey on self-supervised learning: Algorithms, applications, and future trends.
\newblock \emph{PAMI}, 2024.

\bibitem[Guo et~al.(2023)Guo, Jia, Zhang, Li, et~al.]{guo2024recontrast}
Jia Guo, Lize Jia, Weihang Zhang, Huiqi Li, et~al.
\newblock Recontrast: Domain-specific anomaly detection via contrastive reconstruction.
\newblock \emph{NeurIPS}, 36, 2023.

\bibitem[Guo et~al.(2025)Guo, Lu, Zhang, and other]{guo2025dinomaly}
Jia Guo, Shuai Lu, Weihang Zhang, and other.
\newblock Dinomaly: The less is more philosophy in multi-class unsupervised anomaly detection.
\newblock In \emph{CVPR}, pages 20405--20415, 2025.

\bibitem[He et~al.(2024{\natexlab{a}})He, Zhang, Chen, et~al.]{he2024diffusion}
Haoyang He, Jiangning Zhang, Hongxu Chen, et~al.
\newblock A diffusion-based framework for multi-class anomaly detection.
\newblock In \emph{AAAI}, pages 8472--8480, 2024{\natexlab{a}}.

\bibitem[He et~al.(2025)He, Bai, Zhang, et~al.]{he2024mambaad}
Haoyang He, Yuhu Bai, Jiangning Zhang, et~al.
\newblock {MambaAD: Exploring state space models for multi-class unsupervised anomaly detection}.
\newblock \emph{NeurIPS}, 37:\penalty0 71162--71187, 2025.

\bibitem[He et~al.(2016)He, Zhang, Ren, and Sun]{he2016deep}
Kaiming He, Xiangyu Zhang, Shaoqing Ren, and Jian Sun.
\newblock Deep residual learning for image recognition.
\newblock In \emph{CVPR}, pages 770--778, 2016.

\bibitem[He et~al.(2024{\natexlab{b}})He, Jiang, Peng, et~al.]{he2024learning}
Liren He, Zhengkai Jiang, Jinlong Peng, et~al.
\newblock Learning unified reference representation for unsupervised multi-class anomaly detection.
\newblock In \emph{ECCV}, pages 216--232. Springer, 2024{\natexlab{b}}.

\bibitem[Huang et~al.(2022)Huang, Guan, Jiang, et~al.]{huang2022registration}
Chaoqin Huang, Haoyan Guan, Aofan Jiang, et~al.
\newblock Registration based few-shot anomaly detection.
\newblock In \emph{ECCV}, pages 303--319. Springer, 2022.

\bibitem[Hyun et~al.(2024)Hyun, Kim, Jeon, et~al.]{hyun2024reconpatch}
Jeeho Hyun, Sangyun Kim, Giyoung Jeon, et~al.
\newblock Reconpatch: Contrastive patch representation learning for industrial anomaly detection.
\newblock In \emph{WACV}, pages 2052--2061, 2024.

\bibitem[Jing and Tian(2020)]{jing2020self}
Longlong Jing and Yingli Tian.
\newblock Self-supervised visual feature learning with deep neural networks: A survey.
\newblock \emph{PAMI}, 43\penalty0 (11):\penalty0 4037--4058, 2020.

\bibitem[Khosla et~al.(2020)Khosla, Teterwak, Wang, et~al.]{khosla2020supervised}
Prannay Khosla, Piotr Teterwak, Chen Wang, et~al.
\newblock Supervised contrastive learning.
\newblock \emph{NeurIPS}, 33:\penalty0 18661--18673, 2020.

\bibitem[Lee et~al.(2024)Lee, Kim, Park, Woo, and Ko]{lee2024crad}
Joo~Chan Lee, Taejune Kim, Eunbyung Park, Simon~S. Woo, and Jong~Hwan Ko.
\newblock Continuous memory representation for anomaly detection.
\newblock \emph{ECCV}, 2024.

\bibitem[Li et~al.(2021)Li, Sohn, Yoon, and Pfister]{li2021cutpaste}
Chun-Liang Li, Kihyuk Sohn, Jinsung Yoon, and Tomas Pfister.
\newblock Cutpaste: Self-supervised learning for anomaly detection and localization.
\newblock In \emph{CVPR}, pages 9664--9674, 2021.

\bibitem[Li et~al.(2022)Li, Zhan, Wang, et~al.]{li2022towards}
Wujin Li, Jiawei Zhan, Jinbao Wang, et~al.
\newblock Towards continual adaptation in industrial anomaly detection.
\newblock In \emph{ACM MM}, pages 2871--2880, 2022.

\bibitem[Liang et~al.(2025)Liang, Hu, Huang, et~al.]{10904852}
Yun Liang, Zhiguang Hu, Junjie Huang, et~al.
\newblock Tocoad: Two-stage contrastive learning for industrial anomaly detection.
\newblock \emph{IEEE TIM}, 74:\penalty0 1--9, 2025.

\bibitem[Liao et~al.(2024)Liao, Xu, Nguyen, et~al.]{liao2024coft}
Jingyi Liao, Xun Xu, Manh~Cuong Nguyen, et~al.
\newblock {COFT-AD: COntrastive Fine-Tuning for Few-Shot Anomaly Detection}.
\newblock \emph{TIP}, 2024.

\bibitem[Liu et~al.(2024{\natexlab{a}})Liu, Wu, Nie, et~al.]{liu2024unsupervised}
Jiaqi Liu, Kai Wu, Qiang Nie, et~al.
\newblock Unsupervised continual anomaly detection with contrastively-learned prompt.
\newblock In \emph{AAAI}, pages 3639--3647, 2024{\natexlab{a}}.

\bibitem[Liu et~al.(2023{\natexlab{a}})Liu, Chang, Ma, Shan, and Chen]{liu2023diversity}
Wenrui Liu, Hong Chang, Bingpeng Ma, Shiguang Shan, and Xilin Chen.
\newblock Diversity-measurable anomaly detection.
\newblock In \emph{CVPR}, pages 12147--12156, 2023{\natexlab{a}}.

\bibitem[Liu et~al.(2021)Liu, Zhang, Hou, Mian, Wang, Zhang, and Tang]{liu2021self}
Xiao Liu, Fanjin Zhang, Zhenyu Hou, Li Mian, Zhaoyu Wang, Jing Zhang, and Jie Tang.
\newblock Self-supervised learning: Generative or contrastive.
\newblock \emph{TKDE}, 35\penalty0 (1):\penalty0 857--876, 2021.

\bibitem[Liu et~al.(2024{\natexlab{b}})Liu, Wang, Leng, and Zhang]{liu2024multimodal}
Xinyue Liu, Jianyuan Wang, Biao Leng, and Shuo Zhang.
\newblock Multimodal industrial anomaly detection by crossmodal reverse distillation.
\newblock \emph{preprint arXiv:2412.08949}, 2024{\natexlab{b}}.

\bibitem[Liu et~al.(2023{\natexlab{b}})Liu, Zhou, Xu, and Wang]{liu2023simplenet}
Zhikang Liu, Yiming Zhou, Yuansheng Xu, and Zilei Wang.
\newblock Simplenet: A simple network for image anomaly detection and localization.
\newblock In \emph{CVPR}, pages 20402--20411, 2023{\natexlab{b}}.

\bibitem[Lu et~al.(2023)Lu, Yao, Fu, and Jia]{lu2023removing}
Fanbin Lu, Xufeng Yao, Chi-Wing Fu, and Jiaya Jia.
\newblock Removing anomalies as noises for industrial defect localization.
\newblock In \emph{ICCV}, pages 16166--16175, 2023.

\bibitem[Mishra et~al.(2021)Mishra, Verk, Fornasier, et~al.]{btad_ad}
Pankaj Mishra, Riccardo Verk, Daniele Fornasier, et~al.
\newblock {VT-ADL: A vision transformer network for image anomaly detection and localization}.
\newblock In \emph{International Symposium on Industrial Electronics}, pages 01--06, 2021.

\bibitem[Oord et~al.(2018)Oord, Li, and Vinyals]{oord2018representation}
Aaron van~den Oord, Yazhe Li, and Oriol Vinyals.
\newblock Representation learning with contrastive predictive coding.
\newblock \emph{preprint arXiv:1807.03748}, 2018.

\bibitem[Pang et~al.(2021)Pang, Shen, Cao, and Hengel]{pang2021deep}
Guansong Pang, Chunhua Shen, Longbing Cao, and Anton Van~Den Hengel.
\newblock Deep learning for anomaly detection: A review.
\newblock \emph{ACM computing surveys}, 54\penalty0 (2):\penalty0 1--38, 2021.

\bibitem[Parisi et~al.(2019)Parisi, Kemker, Part, Kanan, and Wermter]{parisi2019continual}
German~I Parisi, Ronald Kemker, Jose~L Part, Christopher Kanan, and Stefan Wermter.
\newblock Continual lifelong learning with neural networks: A review.
\newblock \emph{Neural networks}, 113:\penalty0 54--71, 2019.

\bibitem[Ristea et~al.(2022)Ristea, Madan, Ionescu, et~al.]{ristea2022self}
Nicolae-C{\u{a}}t{\u{a}}lin Ristea, Neelu Madan, Radu~Tudor Ionescu, et~al.
\newblock Self-supervised predictive convolutional attentive block for anomaly detection.
\newblock In \emph{CVPR}, pages 13576--13586, 2022.

\bibitem[Roth et~al.(2022)Roth, Pemula, Zepeda, et~al.]{patchcore}
Karsten Roth, Latha Pemula, Joaquin Zepeda, et~al.
\newblock Towards total recall in industrial anomaly detection.
\newblock In \emph{CVPR}, pages 14318--14328, 2022.

\bibitem[Rudolph et~al.(2023)Rudolph, Wehrbein, Rosenhahn, and Wandt]{rudolph2023asymmetric}
Marco Rudolph, Tom Wehrbein, Bodo Rosenhahn, and Bastian Wandt.
\newblock Asymmetric student-teacher networks for industrial anomaly detection.
\newblock In \emph{WACV}, pages 2592--2602, 2023.

\bibitem[Sohn(2016)]{npairs}
Kihyuk Sohn.
\newblock {Improved Deep Metric Learning with Multi-class N-pair Loss Objective}.
\newblock In \emph{NeurIPS}, 2016.

\bibitem[Su et~al.(2021)Su, You, Wang, Qian, Zhang, and Xu]{su2021bcnet}
Xiu Su, Shan You, Fei Wang, Chen Qian, Changshui Zhang, and Chang Xu.
\newblock Bcnet: Searching for network width with bilaterally coupled network.
\newblock In \emph{CVPR}, pages 2175--2184, 2021.

\bibitem[Su et~al.(2022)Su, You, Xie, Zheng, Wang, Qian, Zhang, Wang, and Xu]{su2022vitas}
Xiu Su, Shan You, Jiyang Xie, Mingkai Zheng, Fei Wang, Chen Qian, Changshui Zhang, Xiaogang Wang, and Chang Xu.
\newblock Vitas: Vision transformer architecture search.
\newblock In \emph{ECCV}, pages 139--157. Springer, 2022.

\bibitem[Tack et~al.(2020)Tack, Mo, Jeong, and Shin]{tack2020csi}
Jihoon Tack, Sangwoo Mo, Jongheon Jeong, and Jinwoo Shin.
\newblock {CSI: Novelty detection via contrastive learning on distributionally shifted instances}.
\newblock \emph{NeurIPS}, 33:\penalty0 11839--11852, 2020.

\bibitem[Tang et~al.(2024)Tang, Lu, Xu, et~al.]{tang2024incremental}
Jiaqi Tang, Hao Lu, Xiaogang Xu, et~al.
\newblock An incremental unified framework for small defect inspection.
\newblock In \emph{ECCV}, 2024.

\bibitem[Tian et~al.(2020)Tian, Sun, Poole, et~al.]{tian2020makes}
Yonglong Tian, Chen Sun, Ben Poole, et~al.
\newblock What makes for good views for contrastive learning?
\newblock \emph{NeurIPS}, 33:\penalty0 6827--6839, 2020.

\bibitem[Tien et~al.(2023)Tien, Nguyen, Tran, et~al.]{rd++}
Tran~Dinh Tien, Anh~Tuan Nguyen, Nguyen~Hoang Tran, et~al.
\newblock Revisiting reverse distillation for anomaly detection.
\newblock In \emph{CVPR}, pages 24511--24520, 2023.

\bibitem[Van~der Maaten and Hinton(2008)]{van2008visualizing}
Laurens Van~der Maaten and Geoffrey Hinton.
\newblock {Visualizing data using t-SNE.}
\newblock \emph{Journal of machine learning research}, 9\penalty0 (11), 2008.

\bibitem[Wang et~al.(2024{\natexlab{a}})Wang, Zhu, Gao, et~al.]{real_ad}
Chengjie Wang, Wenbing Zhu, Bin-Bin Gao, et~al.
\newblock Real-iad: A real-world multi-view dataset for benchmarking versatile industrial anomaly detection.
\newblock In \emph{CVPR}, pages 22883--22892, 2024{\natexlab{a}}.

\bibitem[Wang et~al.(2024{\natexlab{b}})Wang, Zhang, Su, and Zhu]{wang2024comprehensive}
Liyuan Wang, Xingxing Zhang, Hang Su, and Jun Zhu.
\newblock A comprehensive survey of continual learning: theory, method and application.
\newblock \emph{PAMI}, 2024{\natexlab{b}}.

\bibitem[Wang et~al.(2021)Wang, Zhang, Shen, Kong, and Li]{wang2021dense}
Xinlong Wang, Rufeng Zhang, Chunhua Shen, Tao Kong, and Lei Li.
\newblock Dense contrastive learning for self-supervised visual pre-training.
\newblock In \emph{CVPR}, pages 3024--3033, 2021.

\bibitem[Wang et~al.(2023)Wang, Peng, Zhang, Yi, Wang, and Wang]{wang2023multimodal}
Yue Wang, Jinlong Peng, Jiangning Zhang, Ran Yi, Yabiao Wang, and Chengjie Wang.
\newblock Multimodal industrial anomaly detection via hybrid fusion.
\newblock In \emph{CVPR}, pages 8032--8041, 2023.

\bibitem[Xia et~al.(2022)Xia, Pan, Li, He, Ma, Zhang, and Ding]{xia2022gan}
Xuan Xia, Xizhou Pan, Nan Li, Xing He, Lin Ma, Xiaoguang Zhang, and Ning Ding.
\newblock Gan-based anomaly detection: A review.
\newblock \emph{Neurocomputing}, 493:\penalty0 497--535, 2022.

\bibitem[Yao et~al.(2024)Yao, Li, Qian, Wang, and Zhang]{HGAD}
Xincheng Yao, Ruoqi Li, Zefeng Qian, Lu Wang, and Chongyang Zhang.
\newblock Hierarchical gaussian mixture normalizing flows modeling for unified anomaly detection.
\newblock In \emph{ECCV}, 2024.

\bibitem[You et~al.(2022)You, Cui, Shen, Yang, Lu, Zheng, and Le]{uniAD}
Zhiyuan You, Lei Cui, Yujun Shen, Kai Yang, Xin Lu, Yu Zheng, and Xinyi Le.
\newblock A unified model for multi-class anomaly detection.
\newblock \emph{NeurIPS}, 35:\penalty0 4571--4584, 2022.

\bibitem[Zavrtanik et~al.(2021)Zavrtanik, Kristan, and Sko{\v{c}}aj]{zavrtanik2021draem}
Vitjan Zavrtanik, Matej Kristan, and Danijel Sko{\v{c}}aj.
\newblock Draem-a discriminatively trained reconstruction embedding for surface anomaly detection.
\newblock In \emph{ICCV}, pages 8330--8339, 2021.

\bibitem[Zhang et~al.(2023{\natexlab{a}})Zhang, Wu, Wang, Chen, and Jiang]{zhang2023prototypical}
Hui Zhang, Zuxuan Wu, Zheng Wang, Zhineng Chen, and Yu-Gang Jiang.
\newblock Prototypical residual networks for anomaly detection and localization.
\newblock In \emph{CVPR}, pages 16281--16291, 2023{\natexlab{a}}.

\bibitem[Zhang et~al.(2024{\natexlab{a}})Zhang, Wang, Li, et~al.]{zhang2024learning}
Jiangning Zhang, Chengjie Wang, Xiangtai Li, et~al.
\newblock Learning feature inversion for multi-class anomaly detection under general-purpose coco-ad benchmark.
\newblock \emph{arXiv preprint arXiv:2404.10760}, 2024{\natexlab{a}}.

\bibitem[Zhang et~al.(2025)Zhang, Chen, Wang, Wang, Liu, Li, Yang, and Tao]{vitad}
Jiangning Zhang, Xuhai Chen, Yabiao Wang, Chengjie Wang, Yong Liu, Xiangtai Li, Ming-Hsuan Yang, and Dacheng Tao.
\newblock Exploring plain vit reconstruction for multi-class unsupervised anomaly detection.
\newblock \emph{Computer Vision and Image Understanding}, 2025.

\bibitem[Zhang et~al.(2023{\natexlab{b}})Zhang, Li, Li, Huang, Shan, and Chen]{zhang2023destseg}
Xuan Zhang, Shiyu Li, Xi Li, Ping Huang, Jiulong Shan, and Ting Chen.
\newblock Destseg: Segmentation guided denoising student-teacher for anomaly detection.
\newblock In \emph{CVPR}, pages 3914--3923, 2023{\natexlab{b}}.

\bibitem[Zhang et~al.(2024{\natexlab{b}})Zhang, Xu, and Zhou]{zhang2024realnet}
Ximiao Zhang, Min Xu, and Xiuzhuang Zhou.
\newblock Realnet: A feature selection network with realistic synthetic anomaly for anomaly detection.
\newblock In \emph{CVPR}, pages 16699--16708, 2024{\natexlab{b}}.

\bibitem[Zhao(2023)]{zhao2023omnial}
Ying Zhao.
\newblock Omnial: A unified cnn framework for unsupervised anomaly localization.
\newblock In \emph{CVPR}, pages 3924--3933, 2023.

\bibitem[Zou et~al.(2022)Zou, Jeong, Pemula, Zhang, and Dabeer]{visa_ad}
Yang Zou, Jongheon Jeong, Latha Pemula, Dongqing Zhang, and Onkar Dabeer.
\newblock Spot-the-difference self-supervised pre-training for anomaly detection and segmentation.
\newblock In \emph{ECCV}, pages 392--408. Springer, 2022.

\end{thebibliography}
}

\end{document}